\documentclass[letterpaper]{article}
\usepackage{aaai2027}
\usepackage[hyphens]{url}
\usepackage{graphicx}
\usepackage{natbib}
\usepackage{caption}
\usepackage{amsmath}
\usepackage{amsfonts}
\usepackage{booktabs}
\usepackage{subcaption}
\usepackage{cleveref}
\usepackage[table]{xcolor}
\usepackage{fontawesome5}

\definecolor{githubblue}{RGB}{105,108,170}
\definecolor{lightblue}{RGB}{220, 235, 255}

\title{4D-WAM: Infusing Spatiotemporal Awareness into World Action Models through Trajectory Fields}
\author {
   Lishan Yang\textsuperscript{\rm 1,\equalcontrib}, 
   Wenxuan Song\textsuperscript{\rm 2,\(\dagger\)}\equalcontrib, 
   Xi Wang\textsuperscript{4}, 
   Pingyue Sheng\textsuperscript{3}, 
   Zheng Fang\textsuperscript{3},\\
   Ziyang Zhou\textsuperscript{2}, 
   Junjie He\textsuperscript{2}, 
   Haodong Yan\textsuperscript{2}, 
   Jiayi Chen\textsuperscript{2}, 
   Nan Sun\textsuperscript{4}, 
   Qiao Sun\textsuperscript{5},\\ 
   Pengwei Wang\textsuperscript{6},
   Lingqiao Liu\textsuperscript{1}, 
   Yan Wang\textsuperscript{4}, 
   Yuxiang Gao\textsuperscript{3}, 
   Feras Dayoub\textsuperscript{1,\(\ddagger\)}, 
   Haoang Li\textsuperscript{2,\(\ddagger\)}
}
\affiliations{
    \textsuperscript{\rm 1}Adelaide University, 
    \textsuperscript{\rm 2}HKUST(GZ), 
    \textsuperscript{\rm 3}COCO Matrix, 
    \textsuperscript{\rm 4}Tsinghua University, 
    \textsuperscript{\rm 5}Fudan University,
    \textsuperscript{\rm 6}BAAI \\
    \quad
    \noindent\faGithub\ \textbf{Code:} \texttt{github.com/lishanyqy/4DWAM}
}

\nocopyright
\begin{document}


\maketitle

\begingroup
\renewcommand{\thefootnote}{}
\footnotetext{\(\ddagger\) Corresponding Author, \(\dagger\) Project Leader.}
\endgroup

\begin{abstract}
Building on recent advances in world models, World Action Models (WAMs) jointly model video prediction and action generation. 
However, they typically represent videos in 2D pixel space, creating a representation gap with 3D space in which robotic actions are executed.
Recent 3D approaches introduce 3D information, but fail to fully exploit the dynamics of 3D structures. 
In this work, we propose 4D-WAM, a model-agnostic training strategy that injects spatiotemporal knowledge from 3D trajectory fields into WAMs through representation alignment. 
To this end, we introduce two complementary objectives: 1) motion alignment, which aligns temporal feature variations across adjacent frames and encourages the model to build local 4D awareness during training, and 2) destination alignment, which guides the model to infer the final destination from the source frame by minimizing the gap between their attention-like similarity distributions. 
Together, these objectives provide both local motion supervision and long-horizon goal guidance, enabling WAMs to learn trajectory-level spatiotemporal representations. 
Extensive in-distribution and out-of-distribution experiments across different base models demonstrate the model's improvements in spatial understanding, execution precision, robustness, generalization, and versatility. 

%
%
%
\end{abstract}



\section{Introduction}
World Models (WMs) have been viewed as a foundation for embodied intelligence by predicting how the physical
environment evolves under interaction~\cite{mastering_world_model, jepa, bruce2024genie, zhu2026sana}. With the rapid progress of video generative models, this predictive capability has begun to transfer from general video
modeling to embodied domains~\cite{bi2026motus, lingbot-va, cosmospolicy, ud-vla, zhao2026frappe, dreamzero, fastwam}, which are named World Action Models (WAMs).
However, these approaches typically aim to model 2D pixels across video frames, overlooking a fundamental principle: videos are 2D projections of a dynamic 3D world~\cite{glassner}.
Moreover, robotic manipulation is inherently grounded in 3D space. 
Therefore, accurate video generation and action prediction both require a faithful understanding of 3D structure and dynamics.

Recent 3D WMs and WAMs address this limitation by augmenting future prediction with depth maps, surface normals, point clouds, or point maps.
Some prior works~\cite{zhen2025tesseract,huang2026pointworld,xu2026kinema4d,action_images,lee2026mu_0} directly incorporate 3D information as input while simultaneously predicting future scene dynamics. 
Another line of work~\cite{x-wam, li2026wam4d} uses depth maps merely as an auxiliary reconstruction target, aiming to encourage models to learn 3D behavioral patterns from 2D inputs. 

However, these approaches suffer from two key limitations: 1) 
They are confined to per-frame 3D representations, such as depth maps, and overlook the 4D nature of dynamic scenes, where 3D spatial structure and its temporal evolution are inherently coupled~\cite{Minkowski1909raum}. 
By reconstructing 3D geometry independently at each time step, these methods fail to explicitly represent the continuous evolution of 3D structures over time~\cite{traceanything}.
2) Predicting pixel-level depth images may drive the model to focus on low-level reconstruction details, rather than developing high-level 4D understanding, which is more essential for effective action prediction.
Thus, training WAMs to perceive spatiotemporal information requires \textit{1) a compact and fundamental representation and 2) a simple but scalable training recipe.}


\begin{figure*}[t]
    \centering
    \includegraphics[width=0.88\linewidth]{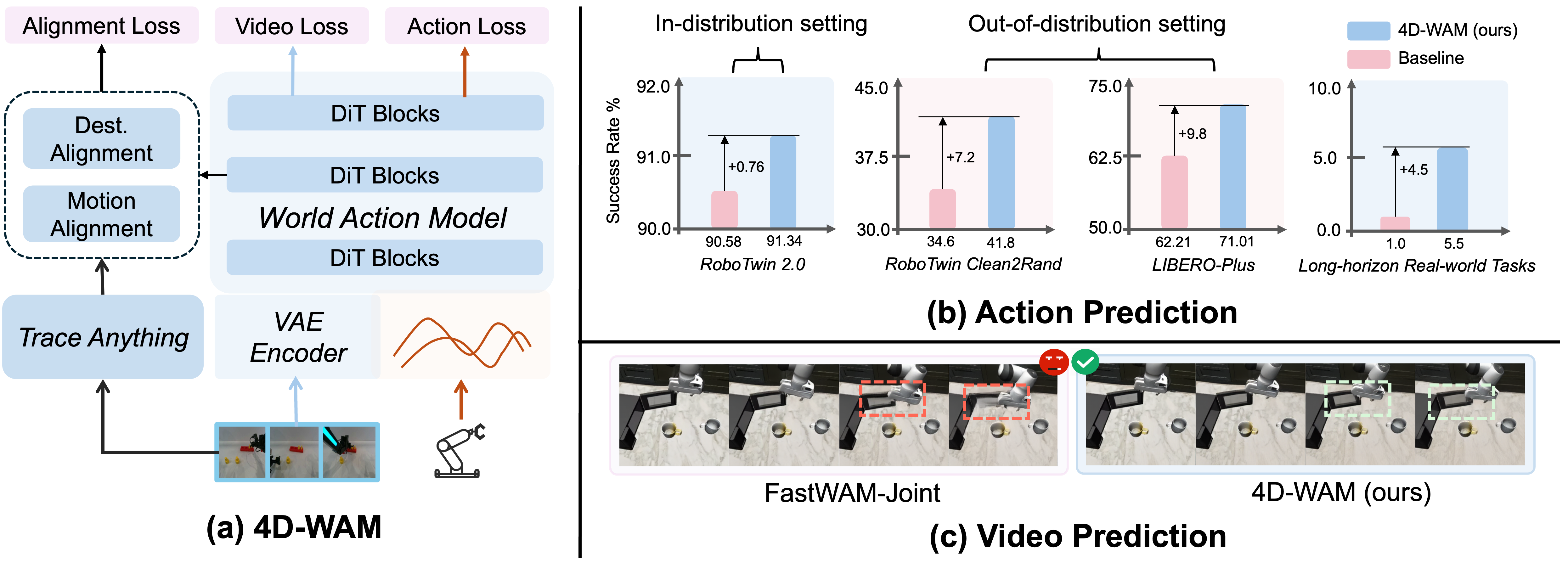}
    \caption{(a) Overview of the 4D-WAM training recipe. 
    (b) 4D-WAM consistently outperforms its corresponding baselines across multiple evaluation settings, with particularly large gains under OOD shifts. (c) Comparison of the video prediction capability. 
    }
    \label{fig:training_pipeline}
\end{figure*}

3D trajectory fields provide such a representation by associating each pixel across frames with a parametric 3D trajectory, thereby directly capturing the motion of objects and robots.
By representing 3D positions as functions of time, these trajectory fields naturally constitute a 4D representation that jointly encodes spatial structure and temporal evolution.
To leverage this 4D representation, we propose 4D-WAM, a training recipe that infuses spatiotemporal awareness into WAM through 3D trajectory fields.
Inspired by representation alignment techniques~\cite{repa, ross, geometryforcing, spatialforcing, song2026reconvla}, 4D-WAM aligns the intermediate visual features of WAM with the 4D features encoded by a 4D foundation model, as shown in \Cref{fig:training_pipeline}.
To align these two representations, our method introduces two complementary alignment objectives: \textbf{Motion Alignment} and \textbf{Destination Alignment}. 
Motion alignment extracts motion-related information from 3D trajectory fields through inter-frame differences and uses cosine similarity to encourage latent 2D world modeling to align with 3D state transition.
%
While motion alignment effectively captures local motion dynamics by enforcing consistency in state transitions, it does not explicitly model the intended destination of manipulated objects. 
To complement this capability, we introduce destination alignment, which encourages the model to associate early moving regions with their corresponding target regions in future frames. 
Specifically, we treat the first frame as the source and the final frame as the destination, extract their representations from the WAM and the 4D foundation model~\cite{traceanything}, respectively, and minimize the discrepancy between the two distributions.
%
This constraint encourages the model to capture target-oriented motion beyond local temporal changes, improving its awareness of manipulation goals.


We conduct extensive in-distribution experiments on LIBERO~\cite{liu2023libero} and RoboTwin 2.0~\cite{chen2025robotwin}, as well as out-of-distribution evaluations on LIBERO-Plus~\cite{liberoplus} and RoboTwin Clean2Rand~\cite{qwenmanip}. 
The results demonstrate that 4D-WAM consistently enhances the model's spatial understanding, execution accuracy, robustness, and generalization capabilities. 
Moreover, its effectiveness is consistently validated when integrated with both FastWAM~\cite{fastwam} and Lingbot-VA~\cite{lingbot-va}, demonstrating its versatility.
We also validate the effectiveness of 4D-WAM in producing accurate video predictions with improved structural consistency.
In summary, the contributions of this paper are threefold:
\begin{itemize}
    \item We propose 4D-WAM, a model-agnostic alignment strategy that injects 3D trajectory fields into WAMs through representation alignment.
    \item We introduce motion alignment and destination alignment to capture temporal dynamics and goal-directed correspondences.
    \item Extensive experiments demonstrate improved precision, robustness, and generalization across different WAM backbones and benchmarks.
\end{itemize}

\section{Related Work}
\subsection{World Action Models}
World Action Models (WAMs) leverage the generative priors of foundation video models together with action experts to support robotic policy learning. UniPi~\cite{unipi} is an early attempt to formulate policy learning as video generation, where the model first generates visual plans and then infers executable actions from the predicted videos.\\ 
\textbf{Inverse Dynamic Models.} Benefiting from the strong large-scale video generation backbones, WAMs based on inverse dynamics models have achieved increasingly promising performance. LAPA~\cite{lapa} learns latent actions from general videos and transfers them to robotic control, while Mimic-Video~\cite{mimic-video} combines a pretrained video backbone with an action decoder to recover executable robot actions from visual dynamics.\\ 
\textbf{Joint Predictions.} Instead of treating video prediction and action generation as two separate stages, these methods learn future world states and robot actions together. Representative works such as UWM~\cite{uwm}, DreamZero~\cite{dreamzero}, and Motus~\cite{bi2026motus} build WAMs on pretrained video diffusion backbones. In this design, video latents learn how the environment changes, while the action branch learns how the robot should act under these changes.\\ 
\textbf{Autoregressive Predictions.} More recent methods use causal attention or autoregressive structures to organize video and action tokens. This allows the model to learn scene evolution and action execution in temporal order. WorldVLA~\cite{cen2025worldvla} unifies VLA and world modeling as an autoregressive action world model. Lingbot-VA~\cite{lingbot-va} further integrates frame prediction and policy execution into a unified autoregressive framework.



\subsection{Spatiotemporal Understanding for WAMs}
%

Early efforts began by spatially grounding robot actions in the visual observation space.
TesserAct~\cite{zhen2025tesseract} subsequently extends world model outputs from 2D videos to 4D scene dynamics by jointly generating RGB, depth, and surface-normal sequences for 4D reconstruction. 
%
PointWorld~\cite{huang2026pointworld} and Kinema4D~\cite{xu2026kinema4d} both convert robot actions into spatially grounded point trajectories. PointWorld uses future 3D robot point flows to predict scene dynamics, while Kinema4D extends this formulation to 4D pointmap sequences and jointly models RGB and geometric world evolution. 
Action Images~\cite{action_images} represents robot controls as decodable multiview action videos, enabling future observations and executable actions to be jointly modeled within a unified video-generation space. 
More recent WAMs further incorporate 4D priors into video-action modeling~\cite{rynn4d,MECo-WAM}.
WAM4D transfers geometric knowledge into transformers through spatial tokens~\cite{li2026wam4d}, while X-WAM introduces RGB-D video prediction and 3D reconstruction to provide explicit spatial supervision~\cite{x-wam}. $\mu_0$ learns an interactive trace world model that predicts future 3D interaction traces and contact regions~\cite{lee2026mu_0}. Although effective, these methods typically introduce additional modalities, prediction heads, reconstruction objectives, or specialized generative pipelines. 
In contrast, we find that DiT-based (Diffusion Transformer) WAMs can acquire 4D representations solely by aligning their intermediate features with motion and source-to-destination trajectory cues.



\section{Methodology}


\subsection{Limitations of Typical Representation Alignment}
\label{sec:limitation_direct_feat_align}

\begin{figure}[t]
    \centering
    \includegraphics[width=\linewidth]{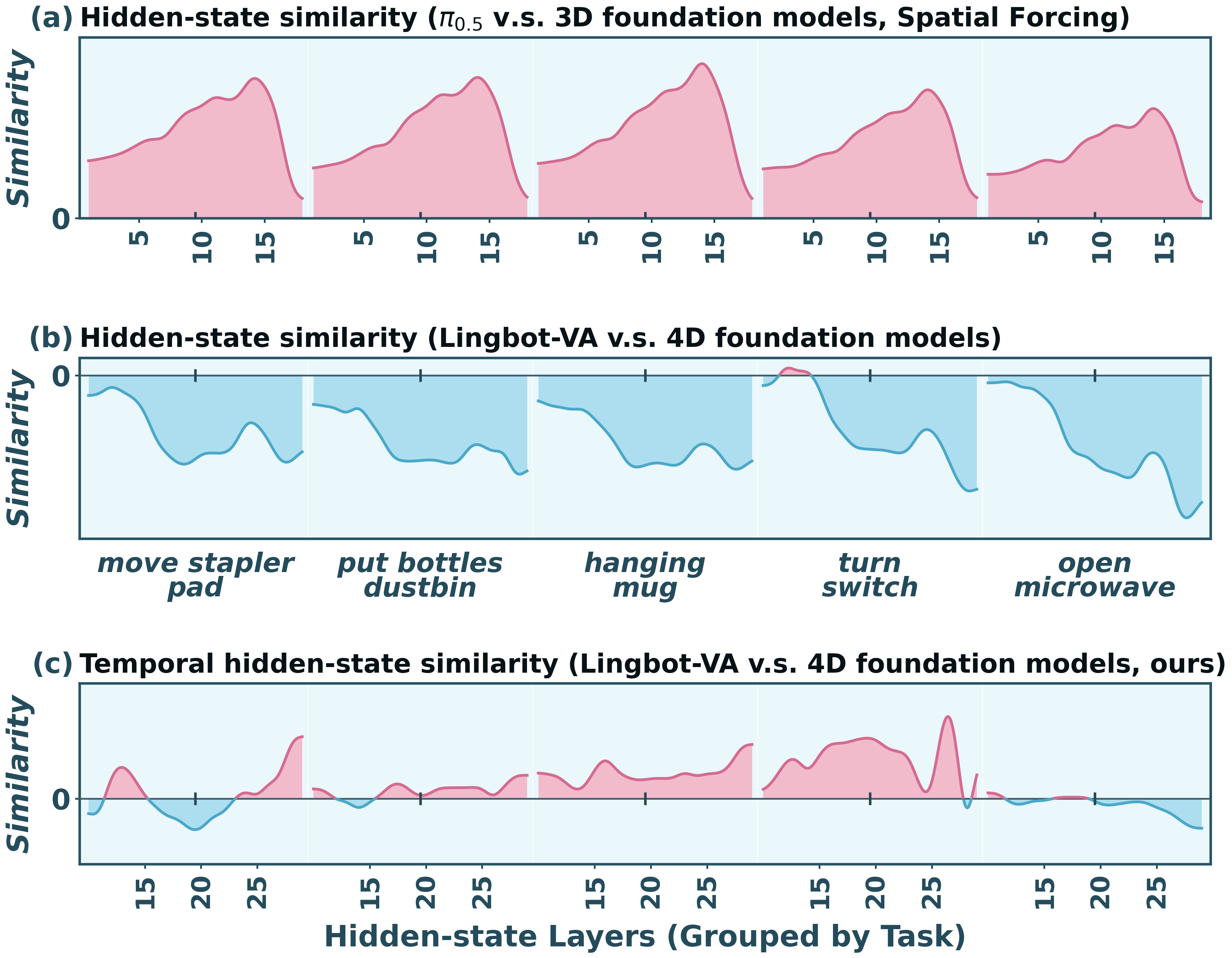}
    \caption{Cosine similarity between different representation spaces across 5 different tasks in RoboTwin 2.0. 
    (a) $\pi_{0.5}$ representations \textit{v.s.} 3D representations from VGGT~\cite{wang2025vggt}. 
    (b) Lingbot-VA representations \textit{vs.} 4D representations from Trace Anything~\cite{traceanything}. 
    (c) Temporal representation differences of Lingbot-VA vs. those of Trace Anything.}
    \label{fig:feature_similarity}
\end{figure}

We first investigate whether the 4D feature alignment strategy, which has proven effective for 3D feature alignment in VLA models~\cite{spatialforcing}, can be directly applied to WAMs. 
However, this straightforward extension consistently leads to inferior performance. 
Figure~\ref{fig:feature_similarity} provides an intuitive explanation for this phenomenon: 
the intermediate representations of a VLA (e.g., $\pi_{0.5}$~\cite{pi05}) exhibit consistently positive similarity to 3D features produced by a foundation model~\cite{wang2025vggt}, whereas the representations of a WAM (e.g., Lingbot-VA~\cite{lingbot-va}) remain uncorrelated or negatively correlated with 4D features~\cite{traceanything} across most layers.
This suggests a substantial discrepancy between the representation spaces of WAMs and 4D foundation models. 
Therefore, instead of aligning absolute features, we ask whether their temporal evolution is more consistent across the two models. To answer this, we compare frame-wise feature differences and observe substantially higher similarity, suggesting that temporal dynamics provide a more suitable supervision signal for alignment.


\subsection{Motion Alignment via Token-wise Frame Differences}
\label{sec:motion_difference_alignment}
As discussed above, directly aligning the intermediate representations of a DiT-based WAM with the features extracted by a reconstruction model may introduce substantial optimization conflicts. Nevertheless, both models are required to capture the temporal evolution of the observed scene. In particular, their representations should encode consistent motion patterns governed by object displacement, spatial interactions, and the physical dynamics of the manipulation process.

\begin{figure*}
    \centering
    \includegraphics[width=0.995\linewidth]{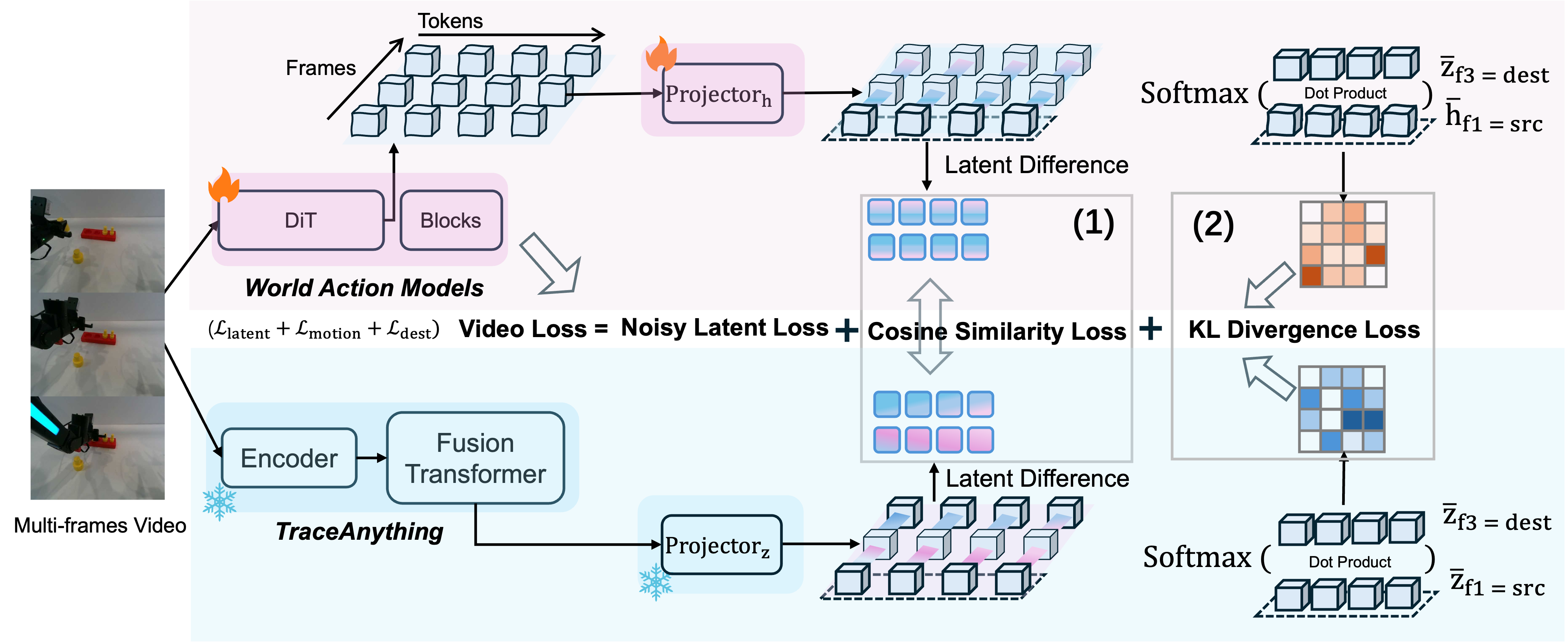}
    \caption{\textbf{Alignment Process of 4D-WAM.} Given a sequence of video frames, 4D-WAM transfers trajectory knowledge from Trace Anything to the WAM through two auxiliary alignment objectives. (1) \textbf{Motion Alignment} projects the two representations into a shared space, computes the change between consecutive frames, and aligns their motion patterns using cosine similarity. (2) \textbf{Destination Alignment} uses the first and final frames to model where the moving regions are expected to arrive, and guides the WAM to learn the same source-to-destination correspondence. 
    }
    \label{fig:alignment_process}
\end{figure*}

Based on this observation, we propose \textbf{Motion Alignment}, which avoids directly matching absolute feature representations and instead aligns their temporal variations across consecutive frames, as shown in~\Cref{fig:alignment_process}. Although the absolute features produced by the two models may reside in different representation spaces, their frame-to-frame changes are expected to describe a consistent underlying motion process. Given an episode containing $F$ observation frames, let 
$\mathbf{H}_{f} \in \mathbb{R}^{N \times d_h}$ 
and 
$\mathbf{Z}_{f} \in \mathbb{R}^{N \times d_z}$ 
denote the intermediate DiT representation and the corresponding 4D reconstruction representation extracted by Trace Anything for the $f$-th frame, respectively. Here, $N$ denotes the number of visual tokens, while $d_h$ and $d_z$ denote their respective feature dimensions. The two input sequences are temporally aligned, such that their token representations $\mathbf{h}_{f,i}$ and $\mathbf{z}_{f,i}$ correspond to the same frame and spatial token, where $f \in \{0,\ldots,F-1\}$ and $i \in \{1,\ldots,N\}$.

    Since the two representations have different feature dimensions and distributions, directly computing their feature differences is inappropriate. We therefore introduce two independent projection networks,
    $\mathcal{P}_{h}$ and $\mathcal{P}_{z}$, to map them into a shared $d$-dimensional feature space. Both projectors are implemented as two-layer multilayer perceptrons:
    \begin{equation}
    \begin{aligned}
        \mathcal{P}_{h}(\mathbf{x})
        &=
        \operatorname{Linear}_{h}^{(2)}
        \left(
            \sigma\left(
                \operatorname{Linear}_{h}^{(1)}(\mathbf{x})
            \right)
        \right), \\
        \mathcal{P}_{z}(\mathbf{x})
        &=
        \operatorname{Linear}_{z}^{(2)}
        \left(
            \sigma\left(
                \operatorname{Linear}_{z}^{(1)}(\mathbf{x})
            \right)
        \right).
    \end{aligned}
    \label{eq:motion_projectors}
    \end{equation}
    
    The projected representations are then obtained as
    \begin{equation}
        \widetilde{\mathbf{h}}_{f,i}
        =
        \mathcal{P}_{h}\left(\mathbf{h}_{f,i}\right),
        \qquad
        \widetilde{\mathbf{z}}_{f,i}
        =
        \mathcal{P}_{z}\left(\mathbf{z}_{f,i}\right),
    \label{eq:projected_motion_features}
    \end{equation}
where $\sigma(\cdot)$ denotes a nonlinear activation function, such as GELU. The projector $\mathcal{P}_z$ is \textbf{frozen} during optimization to preserve a stable target representation space. 

We then characterize the temporal motion of each token using the difference between two consecutive frames. 
Before computing the feature differences, we normalize each token representation in $\ell_2$ norm as $\overline{\mathbf{h}}_{f,i}=\operatorname{Norm}(\widetilde{\mathbf{h}}_{f,i})$ 
and 
$\overline{\mathbf{z}}_{f,i}=\operatorname{Norm}(\widetilde{\mathbf{z}}_{f,i})$.
Specifically, the token-wise differences of the DiT representation and the 4D reconstruction representation are defined as
\begin{equation}
\begin{aligned}
    \Delta \overline{\mathbf{h}}_{f,i} &= \overline{\mathbf{h}}_{f+1,i} - \overline{\mathbf{h}}_{f,i}, \\
    \Delta \overline{\mathbf{z}}_{f,i} &= \overline{\mathbf{z}}_{f+1,i} - \overline{\mathbf{z}}_{f,i},
\end{aligned}
\qquad
f \in \{0,\ldots,F-2\}.
\label{eq:token_motion_difference}
\end{equation}
These difference vectors capture token-wise motion across adjacent frames while reducing reliance on the original feature spaces compared with direct feature alignment. 

We then apply a token-wise cosine distance to align the motion directions in the DiT representations with those encoded by the 4D reconstruction features, 
where $\epsilon$ prevents division by zero. 
For a training batch containing $B$ episodes, the proposed motion alignment loss is formulated as
\begin{equation}
\begin{aligned}
\ell_{\mathrm{motion}}^{\,b,f,i} &= 1-
\frac{
    \left\langle
        \Delta\overline{\mathbf{h}}^{\,b}_{f,i},
        \Delta\overline{\mathbf{z}}^{\,b}_{f,i}
    \right\rangle
}{
    \left\|
        \Delta\overline{\mathbf{h}}^{\,b}_{f,i}
    \right\|_{2}
    \left\|
        \Delta\overline{\mathbf{z}}^{\,b}_{f,i}
    \right\|_{2}
    + \epsilon
},
\\
\mathcal{L}_{\mathrm{motion}}
&=
\frac{1}{B(F-1)N}
\sum_{b=1}^{B}
\sum_{f=0}^{F-2}
\sum_{i=1}^{N}
\ell_{\mathrm{motion}}^{\,b,f,i}.
\end{aligned}
\label{eq:motion_difference_loss}
\end{equation}
Rather than forcing the DiT hidden states to reproduce the absolute representation of Trace Anything, $\mathcal{L}_{\mathrm{motion}}$ only constrains their temporal changes to follow consistent directions. This design alleviates the optimization conflict caused by the discrepancy between the two feature spaces while preserving the original representation capacity of the DiT model. Consequently, the learned hidden states are encouraged to encode temporally coherent and physically meaningful motion patterns, which are particularly important for manipulation tasks requiring accurate and continuous action execution.

\subsection{Source-Conditioned Destination Constraint}
Although motion alignment encourages the model to capture frame-to-frame changes, relying only on local temporal differences may make the learned representation overly sensitive to short-term motion while neglecting the global action target. For manipulation tasks, the model should not only recognize which regions are moving, but also understand where these regions are expected to move in the future. 
To introduce such goal-directed supervision, we propose a Source-Conditioned Destination constraint, which explicitly associates moving source regions at the beginning of a trajectory with their potential destination regions in future frames.

Following notation in Section~\ref{sec:motion_difference_alignment}, $\mathbf{h}$ denotes the extracted hidden representations of the video tokens produced by the DiT backbone, while $\mathbf{z}$ denotes the representations extracted from the teacher model. 
Inspired by the query-key similarity computation in Transformer attention, we treat the tokens from a source frame as queries and those from a destination frame as keys. Let
$\mathbf{h}_{\mathrm{src}}, \mathbf{z}_{\mathrm{src}} \in \mathbb{R}^{N \times d}$
denote the projected WAM and teacher representations of the source frame, respectively, and let
$\mathbf{z}_{\mathrm{dest}} \in \mathbb{R}^{N \times d}$
denote the projected teacher representation of the destination frame.


%
For each source token $i$ and destination token $j$, where $i,j\in\{1,\ldots,N\}$, we compute the student and teacher similarity scores using the normalized projected representations $\overline{\mathbf{h}}$ and $\overline{\mathbf{z}}$:
\begin{equation}
s^{h}_{i,j}=\left\langle\overline{\mathbf{h}}_{\mathrm{src},i},\overline{\mathbf{z}}_{\mathrm{dest},j}\right\rangle,\quad
s^{z}_{i,j}=\left\langle\overline{\mathbf{z}}_{\mathrm{src},i},\overline{\mathbf{z}}_{\mathrm{dest},j}\right\rangle.
\label{eq:source_destination_similarity}
\end{equation}

The corresponding source-conditioned destination distributions are obtained through a temperature-scaled softmax:
\begin{equation}
\begin{aligned}
\log p_h(\cdot\mid i) &=
\operatorname{logsoftmax}\left(\mathbf{s}^{h}_{i}/\tau_{\mathrm{sim}}\right), \\
q_z(\cdot\mid i) &=
\operatorname{softmax}\left(\mathbf{s}^{z}_{i}/\tau_{\mathrm{sim}}\right).
\label{eq:source_destination_distribution}
\end{aligned}
\end{equation}
The similarity vectors are further converted into source-conditioned destination distributions through temperature-scaled normalization. Here,
$p_h(\cdot \mid i)$ denotes the destination probability distribution predicted by the DiT representation for the $i$-th source token, where each element indicates the likelihood that the source token corresponds to a destination token in the future frame. Similarly,
$q_z(\cdot \mid i)$ denotes the corresponding destination distribution estimated by the teacher model, which serves as the supervision target during training. The temperature parameter $\tau_{\mathrm{sim}}$ controls the concentration of both distributions, with a smaller value producing a sharper correspondence distribution.
To transfer the teacher's structural knowledge, we encourage the predicted destination distribution of each source token to match that of the teacher model. Specifically, we minimize the Kullback--Leibler (KL)~\cite{kldiv} divergence between the two distributions:
\begin{equation}
\begin{aligned}
\mathcal{L}_{\mathrm{dest}}
=
\frac{1}{BN}
\sum_{b=1}^{B}
\sum_{i=1}^{N}
D_{\mathrm{KL}}
\left(
q_z^{\,b}(\cdot|i)
\,\Vert\,
p_h^{\,b}(\cdot|i)
\right).
\end{aligned}
\label{eq:destination_loss}
\end{equation}
Unlike direct feature matching, this objective supervises the relational structure between the source and destination frames. 
In our implementation, the first frame is selected as the source frame, while the final frame is used as the destination frame. We regard the final frame as a completion signal that reflects the expected outcome of the manipulation trajectory. In principle, the source and destination can be selected as any two frames separated by a distant temporal interval.

\subsection{Overall Training Objective}
WAMs jointly optimized over video and action branches typically involve two training objectives. 
The latent reconstruction objective,
$\mathcal{L}_{\mathrm{latent}}$, supervises the denoising process in the latent space following the standard diffusion training paradigm. 
Meanwhile, the action prediction objective,
$\mathcal{L}_{\mathrm{action}}$, encourages the model to generate chunks of action that are consistent with the demonstrated manipulation sequence.
Building upon the original training objectives, we further introduce the proposed motion alignment loss 
$\mathcal{L}_{\mathrm{motion}}$ and the source-conditioned Destination Alignment loss 
$\mathcal{L}_{\mathrm{dest}}$, to transfer the temporal motion dynamics and the source to destination correspondence from the teacher model, respectively. The overall training objective is therefore formulated as:
\begin{equation}
\mathcal{L}
=
\mathcal{L}_{\mathrm{latent}}
+
\mathcal{L}_{\mathrm{action}}
+
\lambda_{\mathrm{motion}}
\mathcal{L}_{\mathrm{motion}}
+
\lambda_{\mathrm{dest}}
\mathcal{L}_{\mathrm{dest}},
\label{eq:overall_loss}
\end{equation}
where $\lambda_{\mathrm{motion}}$ and $\lambda_{\mathrm{dest}}$ are balancing coefficients that control the contributions of the two alignment objectives. To avoid overly affecting the learning of video generation, we recommend setting both coefficients below $0.1$. During training, the original WAM objectives remain unchanged, while the proposed alignment losses act as auxiliary supervision to improve the temporal consistency and goal awareness of the learned video representations.

\section{Experiments}



\paragraph{Experimental Setup.}
For all experiments, we perform representation alignment at the 20th layer and conduct experiments using 8 NVIDIA H100 GPUs.
%
For finetuning with FastWAM-Joint as the base model, we followed the official configuration to use a global batch size of 1024 for 30K training steps on RoboTwin 2.0~\cite{chen2025robotwin} and a global batch size of 512 and trained the model for 25K steps on LIBERO~\cite{liu2023libero}. 
For finetuning with Lingbot-VA as the base model in RoboTwin Clean2Rand, we used the clean split of RoboTwin 2.0 as the training set, with a global batch size of 64 for 20K training steps.

\begin{table}[t]
\scriptsize
\centering
\renewcommand{\arraystretch}{0.7}
\setlength{\tabcolsep}{2pt}


\begin{tabular}{lccc@{\hspace{10pt}}ccccc}
\toprule
& \multicolumn{3}{c}{RoboTwin 2.0}
& \multicolumn{5}{c}{LIBERO} \\
\cmidrule(lr){2-4}
\cmidrule(lr){5-9}

\textbf{Model}
& \textbf{Clean}
& \textbf{Random}
& \textbf{Avg.}
& \textbf{Object}
& \textbf{Goal}
& \textbf{Spatial}
& \textbf{Long}
& \textbf{Avg.} \\
\midrule

$\pi_{0.5}$
& 42.98 & 43.84 & 43.41
& 98.2 & 98.0 & 98.8 & 92.4 & 96.9 \\

WorldVLA
& 42.50 & 32.20 & 37.35
& 96.2 & 83.4 & 87.6 & 60.0 & 81.8 \\

X-VLA
& 72.80 & 72.84 & 72.82
& 98.6 & 97.8 & 98.2 & \textbf{97.6} & 98.1 \\

Motus
& 88.66 & 87.02 & 87.84
& \textbf{99.8} & 96.6 & 96.8 & \textbf{97.6} & 97.7 \\

X-WAM
& 89.80 & 90.70 & 90.25
& \multicolumn{5}{c}{--} \\

FastWAM-Joint
& 90.84 & 90.32 & 90.58
& 99.0 & 98.4 & \textbf{99.6} & 95.8 & 98.2 \\

\midrule

\rowcolor{lightblue}
\textbf{4D-WAM (ours)}
& \textbf{91.92}
& \textbf{90.76}
& \textbf{91.34}
& 99.5
& \textbf{98.6}
& 98.8
& 97.4
& \textbf{98.6} \\
\bottomrule
\end{tabular}%
\caption{Success rates (\%) comparison on RoboTwin 2.0 and LIBERO.
Our base model is FastWAM-Joint.
Best results are shown in bold.
}
\label{tab:main_table}
\end{table}













\subsection{In-distribution Experiments}

We evaluate our method on the RoboTwin 2.0 and LIBERO benchmarks. For RoboTwin 2.0, the training set consists of 500 randomized demonstrations and 50 clean demonstrations per task across 50 tasks. For LIBERO, we follow the standard evaluation protocol over four suites (Object, Goal, Spatial, and Long), comprising 40 tasks, with 100 demonstrations available for each task.

In these experiments, 4D-WAM is built upon FastWAM-Joint, which achieves the strongest overall performance among the FastWAM variants on both benchmarks. As shown in Table~\ref{tab:main_table}, 4D-WAM outperforms representative baselines, including $\pi_{0.5}$ and Motus, across the 50 RoboTwin 2.0 tasks. 
Compared with base model FastWAM-Joint, our method achieves absolute success-rate improvements of 1.08\% and 0.44\% under the clean and randomized settings, respectively. 
On LIBERO, where the performance of existing methods is already close to saturation, 4D-WAM further increases the average success rate from 98.2\% to 98.6\%. Improvements are observed on the Object, Goal, and Long suites, demonstrating that the proposed alignment objectives can still provide consistent benefits to a strong WAM baseline.

\begin{table}[t]
\centering
\renewcommand{\arraystretch}{1.12}
\setlength{\tabcolsep}{5pt}

\resizebox{1\columnwidth}{!}{%
\begin{tabular}{lccccccc|c}
\toprule
\textbf{Model} & \textbf{Camera} & \textbf{Robot} & \textbf{Language} & \textbf{Light} & \textbf{Background} & \textbf{Noise} & \textbf{Layout} & \textbf{Average} \\
\midrule
FastWAM-Joint & 27.89 & 60.90 & 89.26 & 86.73 & 49.26 & 51.09 & 74.82 & 62.21 \\
\rowcolor{lightblue}
4D-WAM (ours) & \textbf{45.15} & \textbf{64.26} & \textbf{90.63} & \textbf{94.29} & \textbf{57.71} & \textbf{69.08} & \textbf{79.21} & \textbf{71.01} \\
\bottomrule
\end{tabular}%
}
\caption{\textbf{Success rates (\%) on LIBERO-Plus.} Both models were trained on LIBERO only, while the evaluation is conducted under seven perturbations. Our base model is FastWAM-Joint. Best results are shown in bold.}
\label{tab:liberoplus}
\end{table}


\subsection{Out-of-Distribution (OOD) Experiments}
In these experiments, we evaluate the OOD generalization capability of 4D-WAM on two benchmarks, RoboTwin Clean2Rand~\cite{qwenmanip} and LIBERO-Plus~\cite{liberoplus}. 
In LIBERO-Plus, the model is trained on the clean LIBERO dataset and evaluated with multiple dimensions of unseen variations.
In RoboTwin Clean2Rand, the model is trained on only 50 clean demonstrations per task and evaluated under domain randomization. 

As shown in Table~\ref{tab:liberoplus}, 4D-WAM outperforms FastWAM-Joint under all seven perturbation categories and improves the average success rate by 8.8 percentage point.
This contrasts with the 0.4\% improvement observed on the standard LIBERO benchmark, indicating that the benefits of 4D-WAM are particularly pronounced in terms of robustness and generalization.
Specifically, our model achieves a 17.26-point gain under camera perturbations, demonstrating substantially enhanced spatial understanding.
Under lighting, noise, and background variations, 4D-WAM also reaches high success rates and huge improvements, which further demonstrate the generalization to unseen environmental variations.

Table~\ref{tab:clean2rand} shows, on RoboTwin Clean2rand, all baselines exhibit varying degrees of performance degradation when transferred from the clean setting to the randomized setting. 
StarVLA and FastWAM suffer the most substantial drops, with their randomized-setting success rates falling below 20\%. 
In comparison, $\pi_{0.5}$ demonstrates relatively stronger generalization. 
Lingbot-VA achieves a success rate above 80\% under the Clean setting, but its performance decreases considerably under randomized evaluation. 
To evaluate the versatility and ensure basic performance in the randomized setting, we incorporate our proposed alignment strategy into Lingbot-VA instead of previous FastWAM.
We find that 4D-WAM becomes the only model with an average success rate above 60\% across the two settings. In particular, it substantially improves the randomized performance over the original Lingbot-VA, demonstrating stronger generalization to unseen object configurations and scene variations.

\begin{table}[t]
\centering
\renewcommand{\arraystretch}{1.08}
\setlength{\tabcolsep}{3.5pt}

\resizebox{\columnwidth}{!}{%
\begin{tabular}{
    l
    cccccc
    >{\columncolor{lightblue}}c
}
\toprule
\textbf{Split}
& \textbf{StarVLA}
& \textbf{GR00T-N1.7}
& $\boldsymbol{\pi_{0.5}}$
& \textbf{Abot-M0}
& \textbf{FastWAM}
& \textbf{Lingbot-VA}
& \textbf{4D-WAM (ours)} \\
\midrule

\textbf{Clean}
& 58.1
& 43.6
& 68.5
& 70.7
& 70.2
& \underline{80.7}
& \textbf{81.5} 
\\

\textbf{Rand.}
& 10.6
& 20.7
& \textbf{46.0}
& 36.0
& 1.2
& 34.6
& \underline{41.8} \\
\midrule
\textbf{Average}
& 34.3
& 32.2
& 57.3
& 53.4
& 35.7
& \underline{57.7}
& \textbf{61.7} 
\\

\bottomrule
\end{tabular}%
}
\caption{\textbf{Success rates (\%) on RoboTwin 2.0.}
All models are trained
on the clean split and evaluated on the split with domain randomization.
Our base model is Lingbot-VA.
Best and second-best results are shown in bold and underlined, respectively.}
\label{tab:clean2rand}
\end{table}

\begin{table}[t]
\centering
\footnotesize

\renewcommand{\arraystretch}{0.8}
\setlength{\tabcolsep}{2pt}

\begin{tabular}{lcc}
\toprule
\textbf{Model} & \textbf{PSNR} $\uparrow$ & \textbf{SSIM} $\uparrow$ \\
\midrule
FastWAM-Joint & 19.22 & 0.75 \\
\rowcolor{lightblue} 4D-WAM (ours) & \textbf{21.44} & \textbf{0.83} \\
\bottomrule
\end{tabular}%
\caption{Video prediction quality computed over 100 episodes for each perturbation type in LIBERO-Plus. 
$\uparrow$ denotes higher values are better.
}
\label{tab:video_pred}

\end{table}

\begin{figure}[t]
    \centering
    \includegraphics[width=1.0\linewidth]{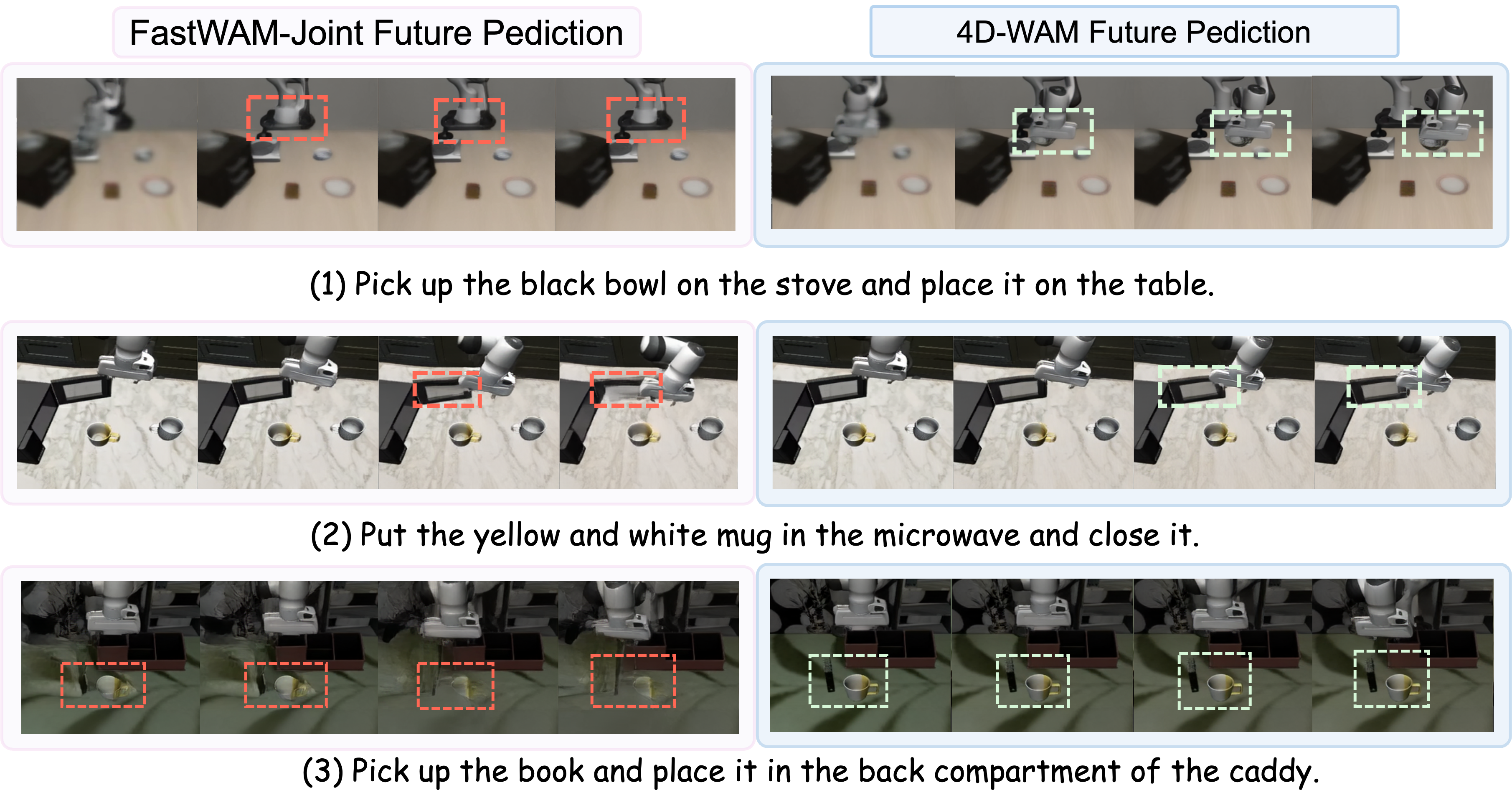}

    \caption{
     \textbf{Video prediction visualization of FastWAM-Joint and our 4D-WAM on LIBERO-Plus. }
     From left to right, the columns show four consecutive future frames. Red boxes indicate poor predictions, while green boxes highlight high-quality generation results.
    }
    \label{fig:video_prediction}
\end{figure}

\subsection{Experiements on Video Prediction}


%

\paragraph{Quantitative Results.} 
To evaluate whether the injected 4D trajectory representation improves future-state modeling, we execute the policies in simulation and compare the predicted observations with the ground-truth observations rendered by the simulator. 
The experiments are conducted on LIBERO-Plus to evaluate the visual prediction capability under OOD settings.
We report PSNR and SSIM, two commonly used metrics for video prediction. PSNR measures pixel-level reconstruction fidelity, whereas SSIM evaluates structural similarity between predicted and ground-truth frames.

Table~\ref{tab:video_pred} shows that incorporating the 4D trajectory field representation substantially improves video prediction quality. Specifically, the PSNR increases from 19.22 to 21.44,and the SSIM increases from 0.75 to 0.83. 
These results indicate that 4D-WAM produces more accurate and structurally consistent future observations than FastWAM-Joint.

\paragraph{Qualitative Results.} 
To provide a more intuitive comparison, we conduct visualized comparisons of the video prediction in Figure~\ref{fig:video_prediction}. 
 
Compared with FastWAM-Joint, 4D-WAM better preserves object motion, target regions, and temporal consistency in future frames. 
This indicates that the injected 4D representation enables the model to capture more structured spatiotemporal transitions, particularly in manipulation-relevant regions, thereby facilitating more precise spatial grounding of actions. 
In the first rollout, FastWAM-Joint fails to preserve the robotic arm from the second frame onward due to camera blur. In the second row, the third frame shows that, unlike FastWAM-Joint, 4D-WAM maintains consistent tracking of the microwave door despite occlusion. In the third row, while grasping the book, 4D-WAM consistently preserves the book’s position from its initial frame to its destination, whereas FastWAM-Joint fails to do so. Our visualizations are consistent with both our methodological design and the observed performance improvements.

Together, these results show that, under OOD conditions, the injected 4D trajectory representation improves the tracking of dynamic and occluded objects, leading to more reliable future-state prediction and more accurate action generation.

\subsection{Ablation Study}
In this section, we present ablation studies on the proposed alignment modules, namely motion alignment (Equation~\ref{eq:motion_difference_loss}) and destination alignment (Equation~\ref{eq:destination_loss}). 
Moreover, we evaluate different alignment layers to assess their impact on performance, motivated by prior findings that intermediate layers are most effective for representation alignment~\cite{repa, spatialforcing}.
The experiments follow the same evaluation setting as Table~\ref{tab:main_table}. 

Table~\ref{tab:ablation_study} shows the results that aligning both objectives at the 20th layer achieves the best performance among the evaluated settings. Removing either alignment objective leads to a performance drop, with both ablated variants achieving a similar accuracy of 98.0\%. Changing the alignment layer also affects performance. Alignment at shallower layers results in accuracy below 98\%, possibly because these layers primarily encode low-level visual information. In contrast, deeper layers tend to capture more task-relevant semantic and reconstruction features. Aligning at or beyond the 20th layer consistently achieves performance above 98.5\%, suggesting that this depth provides a more suitable representation space for our alignment objectives.

\begin{table}[t]
\centering
\footnotesize
\renewcommand{\arraystretch}{0.7}
\setlength{\tabcolsep}{2pt}
\begin{tabular}{lcccc|c}
\toprule
\textbf{Setting} & \textbf{Object} & \textbf{Goal} & \textbf{Spatial} & \textbf{Long} & \textbf{Average} \\
\midrule
\multicolumn{6}{l}{\textit{Alignment layer}} \\
Layer 16 & 99.8 & 97.0 & 98.2 & 96.8 & 97.9 \\
Layer 18 & 99.2 & 97.6 & 98.0 & 94.8 & 97.4 \\
\rowcolor{lightblue} \textbf{Layer 20 (ours)} & 99.6 & 98.6 & 98.8 & 97.4 & \textbf{98.6} \\
Layer 22 & 100.0 & 98.6 & 98.8 & 96.6 & 98.5 \\
Layer 24 & 99.0 & 98.4 & 98.0 & 95.8 & 97.8 \\
\midrule
\multicolumn{6}{l}{\textit{Training objective}} \\
w/o Destination & 99.4 & 97.6 & 98.4 & 96.6 & 98.0 \\
w/o Motion & 99.8 & 97.2 & 99.0 & 96.0 & 98.0 \\
\bottomrule
\end{tabular}
\caption{Ablation study of alignment layer and training objectives in 4D-WAM based on FastWAM on LIBERO.}
\label{tab:ablation_study}
\end{table}




\begin{table}[t]

\centering

\resizebox{\columnwidth}{!}{%

\begin{tabular}{lcccccccc|cc}

\toprule

Model
& \multicolumn{2}{c}{\shortstack{Cylinder \\ Insertion}}
& \multicolumn{2}{c}{\shortstack{Towel \\ Flattening}}
& \multicolumn{2}{c}{\shortstack{Toolbox Wrench \\ Localization}}
& \multicolumn{2}{c}{\shortstack{Rotating Light \\ Pick-and-place}}
& \multicolumn{2}{c}{Average} \\
\cmidrule(lr){2-3}
\cmidrule(lr){4-5}
\cmidrule(lr){6-7}
\cmidrule(lr){8-9}
\cmidrule(lr){10-11}
& Progress & SR
& Progress & SR
& Progress & SR
& Progress & SR
& Progress & SR \\

\midrule

Lingbot-VA & 28.1 & 0.0 & 2.5 & 0.0 & 6.7 & 0.0 & 25.0 & 4.0 & 20.3 & 1.0\\

\rowcolor{lightblue}

4D-WAM (ours) 
& 50.6 & 2.0
& 7.5 & 4.0 
& 8.3 & 12.0 
& 23.8 & 4.0
& \textbf{31.8}& \textbf{5.5} \\

\bottomrule

\end{tabular}%

}
\caption{\textbf{Results of 4 real-world experiments.} Each task is evaluated through 50 trials.}
\label{tab:real}
\end{table}

\begin{figure}[t]
    \centering
    \includegraphics[width=1.0\linewidth]{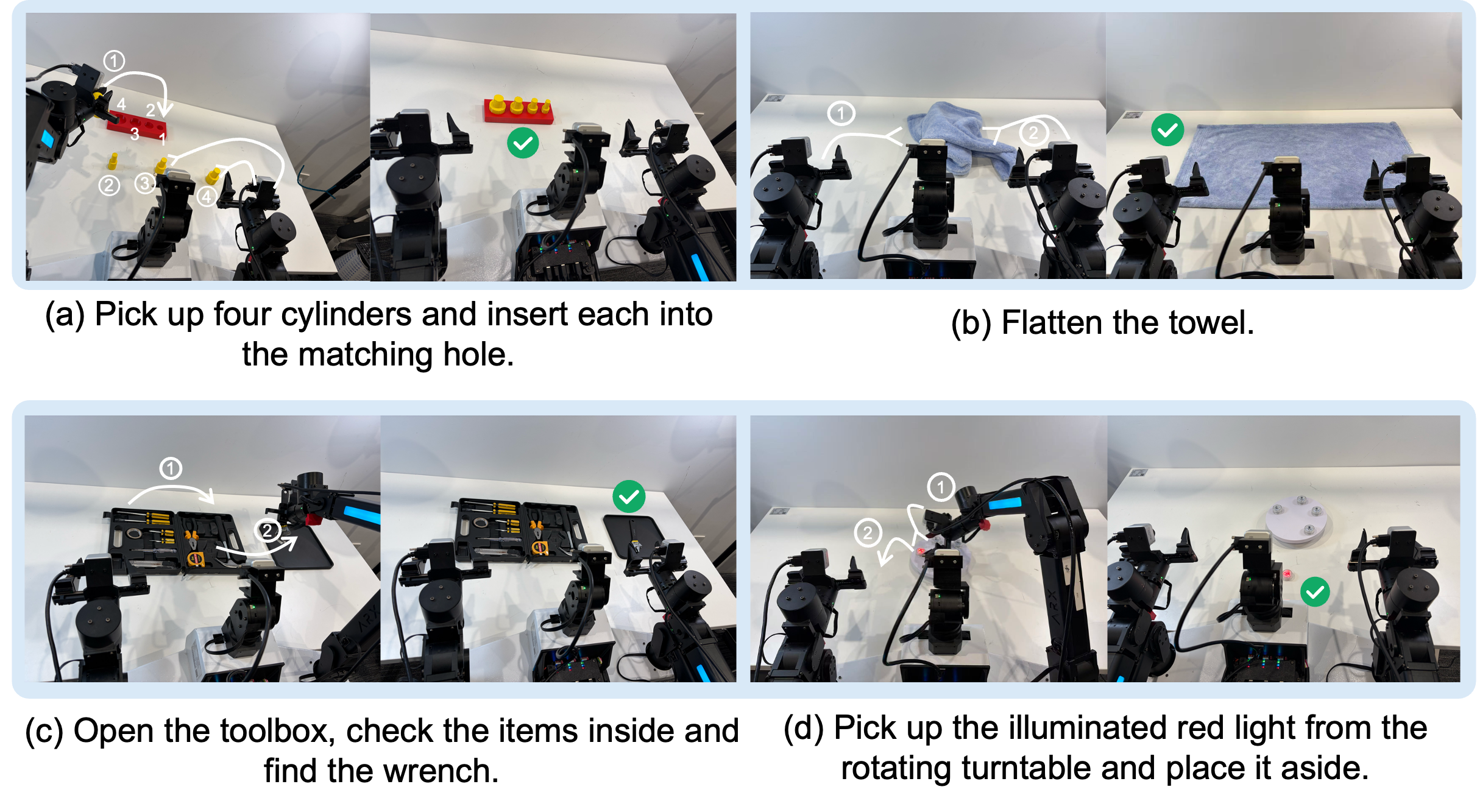}
    \caption{Visualization of 4 long-horizon real-world tasks.
    }
    \label{fig:real_setup}
\end{figure}

\subsection{Real-world Experiments}
As shown in Figure~\ref{fig:real_setup}, we conduct real-world experiments using the ARX LIFT2 bimanual manipulator. Each arm comprises a 6-DoF manipulator and a 1-DoF gripper. 
The platform is equipped with one primary camera and two wrist-mounted cameras. 
We design four tasks covering diverse dimensions of spatiotemporal capability: (a) inserting four randomly placed cylinders into their size-matched holes, which evaluates fine-grained manipulation accuracy and long-horizon manipulation capability;
(b) flattening the towel, which assesses the capability of manipulating deformable objects;
(c) opening a toolbox, inspecting its contents sequentially, and locating the wrench, which evaluates spatial reasoning and long-horizon manipulation capability;
and (d) picking up the illuminated red light from a rotating turntable and placing it aside, which assesses the understanding of dynamics and spatiotemporal reasoning.
We collect 550 demonstrations across the four tasks. 
We use Lingbot-VA as the base model because it is pretrained on large-scale robot data. 
Both 4D-WAM and Lingbot-VA are jointly trained on all four tasks. 
We evaluate models in 50 trials per task. We report both the success rate (SR) and progress as the evaluation metrics. Progress indicates the proportion of subtasks completed.

Table~\ref{tab:real} reports the real-world performance of 4D-WAM and Lingbot-VA across four tasks. 4D-WAM outperforms Lingbot-VA on all tasks. 
Due to error accumulation over long horizons, Lingbot-VA struggles to complete three of the four tasks. 
In contrast, we observe that, owing to its improved prediction precision, 4D-WAM can complete these tasks successfully with 5.5\% total success rates.
In particular, it improves the sub-task progress by more than 10\% on average. 
%
These results suggest that 4D-WAM improves the model’s spatiotemporal understanding and long-horizon manipulation capabilities, while its effectiveness generalizes to real-world settings.

\section{Conclusion}
In this work, we introduce a model-agnostic alignment strategy, 4D-WAM, for improving spatiotemporal perception in WAMs. 
The proposed motion and destination alignment objectives guide the model to capture temporal dynamics and identify episode-level target regions from 3D trajectory fields. 
Across multiple benchmarks, 4D-WAM achieves competitive manipulation performance with clear gains in in-distribution, challenging OOD, and real-world settings. 
Analyses of video predictions further show more accurate object motion and stronger spatiotemporal consistency. 
All of them demonstrate improvements in both downstream performance and generalization.

\appendix

\section{Auxiliary Training Cost}
\label{sec:training_cost}
\begin{table}[t]
\centering
\footnotesize
\renewcommand{\arraystretch}{1.0}
\setlength{\tabcolsep}{2pt}

\resizebox{\columnwidth}{!}{
\begin{tabular}{lccc}
\toprule
\textbf{Metric}
& Baseline
& 4D-WAM
& 4D-WAM (Online) \\
\midrule
Time per step (s)
& 7.3 & 7.5 & 8.4 \\

Peak GPU memory (\%)
& 77.3 & 82.7 & 96.0 \\

Training time (8 GPUs, h)
& 51 & 52 & 59 \\

Extra trainable params. (MiB)
& -- & 56 & 56 \\
\bottomrule
\end{tabular}
}
\caption{\textbf{Training cost compared with the baseline.} Time per step denotes the time required to complete one optimization step. Peak GPU memory is the maximum memory utilization observed across eight GPUs. Total training cost is reported in GPU hours using eight GPUs. Additional trainable parameters correspond to those introduced by the alignment modules.}

\label{tab:train_cost}
\end{table}

4D-WAM introduces trajectory supervision only during training. 
The frozen Trace Anything~\cite{traceanything} extracts
4D trajectory representations, which are used to construct the motion
alignment and destination alignment objectives. 
Since the teacher model is not updated, the additional trainable parameters come only from the lightweight projectors. 
Let $C_{\mathrm{WAM}}$ denote the computational cost of the forward and
backward passes of the original WAM, while $C_{\mathrm{traj}}$ and
$C_{\mathrm{align}}$ denote the costs of trajectory feature extraction and
the proposed alignment objectives, respectively. The overall training cost
can be written as
\begin{equation}
C_{\mathrm{4D\text{-}WAM}}
=
C_{\mathrm{WAM}}
+
C_{\mathrm{traj}}
+
C_{\mathrm{align}}.
\end{equation}
The alignment objectives involve only lightweight operations, including token-wise feature differences, cosine similarities, and source-to-destination similarity computation. Therefore,
$C_{\mathrm{align}}$ is relatively small, and the primary additional cost comes from trajectory feature extraction, i.e., $C_{\mathrm{traj}}$. 
To reduce this overhead, we pre-extract and cache the trajectory representations for all training sequences. We refer to this setting as 4D-WAM, while 4D-WAM (Online) extracts the trajectory features during each training iteration. 
Table~\ref{tab:train_cost} compares the computational costs of these two settings with the original FastWAM-Joint baseline. All models are trained on LIBERO for 25k steps with a global batch size of 512 using eight NVIDIA H100 80\,GiB GPUs.

Compared with the baseline, cached 4D-WAM increases the iteration time from 7.3\,s to 7.5\,s and the total training cost from 51 to 52 eight GPU hours. 
Its peak GPU memory usage increases by 5.4 percentage points. In contrast, online trajectory extraction requires 59 GPU hours and raises peak GPU memory usage to 96.0\%. 
These results show that pre-extracting trajectory representations substantially reduces the training overhead, leaving only a small computational gap between 4D-WAM and the baseline. The alignment projectors introduce only 56\,MiB of additional trainable parameter storage.
After training, Trace Anything, the alignment projectors, and all alignment objectives are removed. Consequently, 4D-WAM retains exactly the same inference architecture and parameter count as the original WAM, introducing no additional inference latency or memory consumption.

\begin{figure}
    \centering
    \includegraphics[width=1.0\linewidth]
    {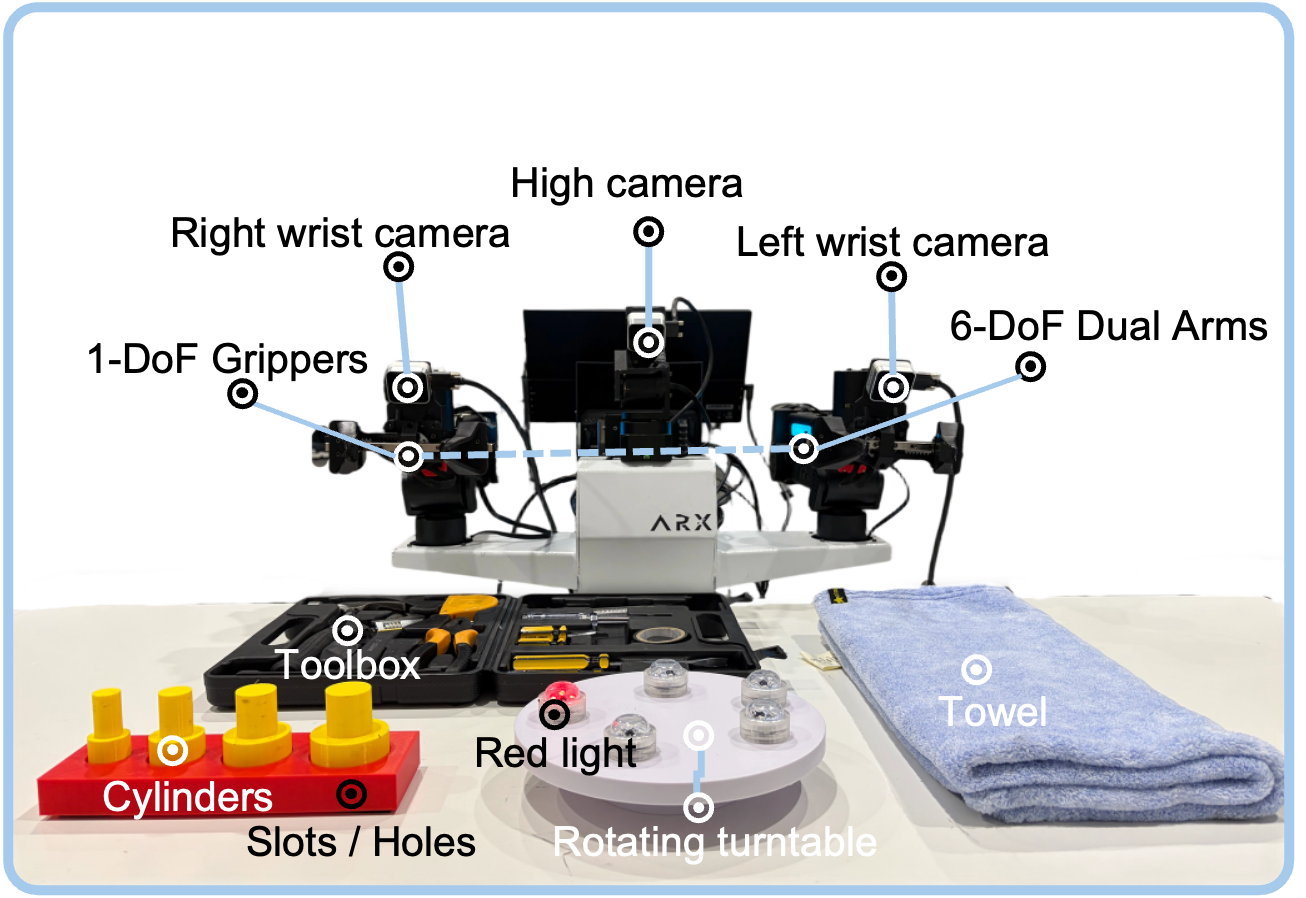}
    \caption{Overview of the real-world robotic platform and manipulated objects.}
    \label{fig:overview_real}
\end{figure}

\begin{figure*}[t]
    \centering
    \includegraphics[width=\textwidth]
    {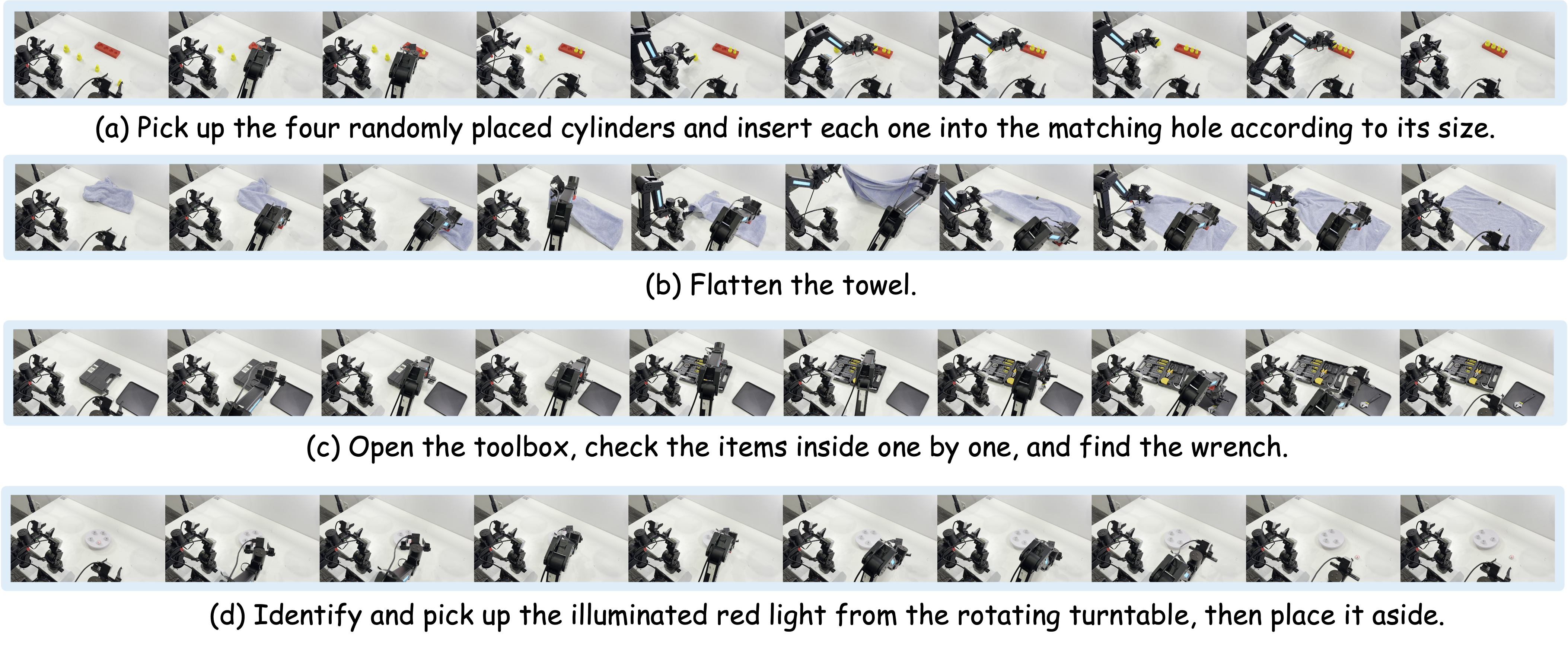}
    \caption{For each of the four tasks, the complete manipulation process is shown sequentially from left to right.}
    \label{fig:sequence_4tasks}
\end{figure*}

\section{Real-world Task and Dataset}
\label{sec:real_details}
\textbf{Setup.} We conduct real-world experiments using an ARX LIFT2 bimanual manipulator. As shown in Figure~\ref{fig:overview_real}, the robot consists of two 6-DoF arms, each equipped with a 1-DoF gripper. Three RGB cameras are used to capture the scene from complementary viewpoints: one is positioned between the two arms, while the other two are positioned near the left and right arms, respectively.\\ 
\textbf{Data Collection.} The main objects used in our experiments are shown in Figure~\ref{fig:overview_real}, including a toolbox, four cylinders of different sizes, a rotating turntable with lights, and a towel. For each task, we arrange the corresponding objects on the table and collect demonstrations by operating the robot through the complete task sequence. We design four real-world tasks, consisting of two bimanual tasks (a and b) and two single-arm tasks (c and d), as illustrated in Figure~\ref{fig:sequence_4tasks}. 
We collect 550 in total. Specifically, we collect 100 episodes for task (a), 150 for task (b), 200 for task (c), and 100 for task (d).
\\
%
\textbf{Real-world Tasks.} Figure~\ref{fig:sequence_4tasks} plays four tasks designed frame by frame. (a) Four cylinders of different sizes are randomly arranged in front of their corresponding slots, which are placed at the center of the table. Each arm is responsible for two cylinders. The robot must identify the cylinders by size and insert them into the corresponding slots in ascending order until all four cylinders are placed correctly. 
(b) A crumpled towel is initially placed on the table. The two arms grasp opposite ends of the towel, pull it open, and place it flat on the tabletop.
(c) A tray is placed next to a closed toolbox. The right arm opens the toolbox, locates and grasps the wrench inside, and places it in the tray.
(d) Several lights are placed on a running turntable. While the turntable is rotating, the robot must locate the red light, grasp it, and place it on the table. 
\begin{table}[t]
\centering

\resizebox{\columnwidth}{!}{
\begin{tabular}{lcccccc}
\toprule
Model & KL $\downarrow$ & CE $\downarrow$ & Top-1 $\uparrow$ & Top-5 $\uparrow$ & Top-10 $\uparrow$ & Rank $\downarrow$ \\

\midrule
\multicolumn{7}{c}{\textit{Frame-to-Frame Position Prediction (LIBERO)}}\\
\midrule
FastWAM-Joint* & 0.1474 & 4.1376 & 0.4000 & 0.7678 & \underline{0.8781} & 5.28 \\
FastWAM-Joint  & \underline{0.1470} & \underline{4.1372} & \underline{0.4037} & \underline{0.7747} & 0.8759 & 5.28 \\
\rowcolor{lightblue}
4D-WAM (ours)    & \textbf{0.1396} & \textbf{4.1298} & \textbf{0.4197} & \textbf{0.7872} & \textbf{0.8853} & \textbf{4.98} \\
\midrule

\multicolumn{7}{c}{\textit{Destination Prediction (LIBERO)}}\\
\midrule
FastWAM-Joint*  & \underline{0.0993} & \underline{4.2149} & \underline{0.4344} & \underline{0.8037} & \underline{0.8969} & \underline{4.59} \\
FastWAM-Joint  & 0.1012 & 4.2167 & 0.4150 & 0.8013 & 0.8950 & 4.66 \\
\rowcolor{lightblue}
4D-WAM (ours)     & \textbf{0.0946} & \textbf{4.2102} & \textbf{0.4487} & \textbf{0.8256} & \textbf{0.9113} & \textbf{4.07} \\

\midrule

\multicolumn{7}{c}{\textit{Frame-to-Frame Position Prediction (LIBERO-Plus)}}\\
\midrule
FastWAM-Joint* & \underline{0.2280} & \underline{4.2306} & 0.2981 & 0.6391 & 0.7859 & \underline{8.39} \\
FastWAM-Joint   & 0.2301 & 4.2328 & \underline{0.3022} & \underline{0.6475} & \underline{0.7897} & 8.41 \\
\rowcolor{lightblue}
4D-WAM (ours)       & \textbf{0.2237} & \textbf{4.2264} & \textbf{0.3159} & \textbf{0.6594} & \textbf{0.7978} & \textbf{8.07} \\

\midrule
\multicolumn{7}{c}{\textit{Destination Prediction (LIBERO-Plus)}}\\
\midrule
FastWAM-Joint* & 0.2121 & 4.3600 & \underline{0.2544} & \underline{0.5631} & \underline{0.7025} & 11.56 \\ 
FastWAM-Joint  & \underline{0.2108} & \underline{4.3587} & 0.2487 & 0.5519 & 0.6944 & \underline{11.38} \\

\rowcolor{lightblue}
4D-WAM (ours)      & \textbf{0.2074} & \textbf{4.3553} & \textbf{0.2625} & \textbf{0.5694} & \textbf{0.7056} & \textbf{10.91} \\

\bottomrule
\end{tabular}
}
\caption{\textbf{Trajectory prediction probing on LIBERO and LIBERO-Plus.} Each encoder is frozen and only a linear probe is trained. Best results within each task are shown in bold. FastWAM-Joint* represents the untrained model.}

\label{tab:trajectory-probe}
\end{table}
\section{Probing Experiments: Mechanism of 4D Awareness}
\label{sec:probing_exp}

To examine whether 4D-WAM learns more trajectory-aware representations, we conduct a frozen-representation probing experiment on LIBERO~\cite{liberoplus} and LIBERO-Plus~\cite{liberoplus}. For each model, we freeze the WAM backbone and extract visual-token representations from the 20th DiT block, which is the same layer used for trajectory field alignment. The extracted representations are then fed into two independent lightweight MLP probes. We generate the predictions from two probes for:
\begin{itemize}
\item Frame-to-frame position prediction (F2F): input tokens are matched from one latent frame to the next.
\item Source-to-destination correspondence (Dest.): input tokens in the first latent frame are matched to tokens in the final latent frame.
\end{itemize}
First, in frame-to-frame position prediction, the probe takes the representation of token $i$ at frame $t$ and directly predicts the position of its corresponding token in the next frame $t+1$. This evaluates whether local temporal motion information is encoded in the representation. Second, in destination prediction, the probe takes the representation of a source token in the first frame and predicts its destination position in the final frame. This evaluates whether the representation captures goal-directed trajectory information. For both tasks, Trace Anything representations provide the target trajectory field for alignment, and the probe outputs a score distribution over candidate token positions. We report KL divergence~\cite{kldiv} and cross-entropy (CE)~\cite{cross_entropy} between the probe prediction and the Trace Anything target distribution. Top-$k$ accuracy measures whether the aligned target token is ranked among the top $k$ predicted candidates, and Rank reports its average ranking position.

Table~\ref{tab:trajectory-probe} reports the probing results. FastWAM-Joint$^{*}$~\cite{fastwam} denotes the original checkpoint without post-training. Across both LIBERO and LIBERO-Plus, 4D-WAM consistently achieves lower KL, lower CE, higher Top-$k$ accuracy, and lower mean rank than the FastWAM-Joint variants. On LIBERO, 4D-WAM improves frame-to-frame Top-1 accuracy from $0.4037$ to $0.4197$, while reducing the mean rank from $5.28$ to $4.98$. For destination prediction, the gain is larger, with Top-1 accuracy increasing from $0.4150$ to $0.4487$ and mean rank decreasing from $4.66$ to $4.07$. The same trend holds on LIBERO-Plus, indicating that the trajectory-aware information remains more accessible under distribution shifts. For destination prediction on LIBERO-Plus, 4D-WAM improves Top-1 accuracy from $0.2487$ to $0.2625$ and reduces mean rank from $11.38$ to $10.91$. 
These results suggest that 4D-WAM improves the capability of capturing local dynamics and predicting precise destinations, both of which are essential for coherent rollouts and thereby contribute to improved downstream performance.

%



\bibliography{aaai2027}


\end{document}